%% file: main.tex
\documentclass[10pt,twocolumn,letterpaper]{article}

\usepackage{cvpr}              
\input{preamble}
\definecolor{cvprblue}{rgb}{0.21,0.49,0.74}
\usepackage[pagebackref,breaklinks,colorlinks,allcolors=cvprblue]{hyperref}
\usepackage{amsmath}
\usepackage{algorithm}
\usepackage{algpseudocode}
\usepackage{multirow}
\usepackage{svg}
\usepackage{array}  
\svgpath{{sec/}}
\setsvg{inkscapelatex=false}      
\usepackage{array}
\usepackage{marvosym}
\usepackage{pifont}
\newcommand{\cmark}{\ding{51}} 
\newcommand{\xmark}{\ding{55}} 
\usepackage{makecell}
\def\paperID{***} 
\def\confName{CVPR}
\def\confYear{2026}

\title{Diffusion Forcing Planner: History-Annealed Planning with Time-Dependent Guidance for Autonomous Driving}

\author{
Zehan Zhang$^{1,3}$\thanks{Equal contribution.}   \thanks{Work done during an internship at Yinwang Intelligent Technology Co., Ltd.} \quad
Neng Zhang$^{2}$\footnotemark[1] \quad
Yaoyi Li$^{2}$\thanks{Corresponding author.} \quad
Jia Cai$^{2}$ \quad
Zhiling Wang$^{3}$\footnotemark[3]
\\[0.5em]
$^{1}$University of Science and Technology of China \\
$^{2}$Yinwang Intelligent Technology Co., Ltd \\
$^{3}$Hefei Institutes of Physical Science, Chinese Academy of Sciences
\\[0.3em]  
\tt\small{zhzhang@mail.ustc.edu.cn} \quad  
\tt\small{zhangneng3@huawei.com} \\
\tt\small{\{liyaoyi,caijia\}@yinwang.com}\\
\tt\small{zlwang@hfcas.ac.cn}
}

\begin{document}
\maketitle
\input{sec/0_abstract}    
\input{sec/1_intro}
\input{sec/2_formatting}
\input{sec/3_finalcopy}

\input{sec/4_Experiments}
\input{sec/conclusion}

\input{sec/X_suppl}
{
    \small
    \bibliographystyle{ieeenat_fullname}
    \bibliography{main}
}
\end{document}

%% file: sec/0_abstract.tex
\begin{abstract}
Learning-based motion planners, despite recent progress, often suffer from temporal inconsistency. Small perturbations across frames can accumulate into unstable trajectories, degrading comfort and safety in closed-loop driving. Several methods attempt to inject history as a static conditioning signal to stabilize outputs, only to induce the planner to copy historical patterns instead of adapting to environment contexts. To address this limitation, we propose Diffusion Forcing Planner (DFP), a diffusion-based planning framework driven by history-guided control. Specifically, DFP decomposes the full trajectory into history, current and future segments, and assign independent noise levels to each segment. The model jointly denoises the historical and the future segments, enforcing a heterogeneous joint diffusion process. At inference, classifier-free guidance (CFG) is applied to steer future sampling using annealed history in a controllable manner. Closed-loop evaluation and comprehensive ablations on nuPlan show that DFP achieves competitive performance while producing continuous, stable, and controllable motion plans in complex driving scenarios.
\end{abstract}

%% file: sec/1_intro.tex
\section{Introduction}
Diffusion models\cite{ho2020denoising} have recently shown strong promise for autonomous driving thanks to their ability to represent multi-modal distributions and generate high-quality long-horizon outputs. They are naturally suited to planning tasks that require multi-modal trajectory outputs and long-horizon action generation\cite{chi2025diffusion,black2024pi_0}. Recent work\cite{diffusiondrive, zhao2025diffe2e, wen2024diffusion,black2024pi_0,black2025_pi05} has integrated diffusion models into end-to-end (E2E) pipelines and vision–language–action (VLA) frameworks to improve the human-like driving behavior of autonomous systems. 

Despite the advantages, diffusion-based policies trained via imitation learning remain sensitive to noise in demonstrations and scene contexts\cite{mandlekar2021matters}. Even in seemingly similar environments, the policy can drift substantially in response to small perturbations in the scene contexts, leading to noticeable frame-to-frame instability in trajectories\cite{black2025real}. As a result, improving the temporal consistency and robustness of planned trajectories has become a key open challenge.

To address this challenge, we leverage the historical trajectory to guide the generation of future plans, thereby improving frame-to-frame consistency. From a decision-modeling perspective, motion planning is inherently non-Markovian: a reasonable action depends not only on the current observation, but also on past observations and actions\cite{torne2025learninglongcontextdiffusionpolicies}. However, in complex traffic scenes, if history is treated as a static condition on equal footing with the environment, the model often tends to replicate historical patterns rather than adjust future decisions in response to environment changes\cite{chen2025drivinggpt, de2019causal, li2024ego, cheng2024rethinking}. This motivates a paradigm in which history is used in a controllable way: it should neither be ignored nor imposed as an unconditional hard constraint. We draw inspiration from Diffusion Forcing Transformer (DFoT) in video diffusion\cite{song2025history}, which uses a noising-as-masking mechanism to selectively expose and anneal historical segments, thereby balancing generation quality and stability. 

Building on this insight, we propose Diffusion Forcing Planner (DFP), a history-guided, chunkwise diffusion framework for planning. The core idea is to partition the full trajectory into history, current, and future chunks, and to sample an independent diffusion timestep for each chunk to implement noising-as-masking. During training, we jointly predict both history and future, forcing the model to learn causally consistent conditional generation under a variety of configurations in which history may be available or masked, and different parts of the future may be visible or occluded. Unlike video generation where historical frames are genuine content, outdated motion patterns in driving history can actively mislead current decisions. To address this, we design a history-annealed classifier-free guidance (CFG)\cite{ho2022classifier} scheme at inference time: we run in parallel a history–annealed branch and a history-noise branch, and linearly fuse their outputs with a tunable coefficient, enabling a controllable trade-off between stability and flexibility. To avoid excessive pull from history onto the future, we apply time-dependent annealing to the ground-truth history. It is initialized close to noise and then rapidly returns to the clean signal. The diffusion timesteps of future chunks are annealed independently per chunk, consistent with the chunkwise-independent time design used in training. Each chunk has its own SDE marginal\cite{song2020score} and its own noise level, thereby distinguishing history and future while enhancing the flexibility and continuity of the policy across chunks. Evaluation on the large-scale real-world autonomous planning benchmark \emph{nuPlan}~\cite{caesar2021nuplan} shows that DFP attains state-of-the-art closed-loop performance among learning-based baselines when scored directly on raw model outputs without post-processing, demonstrating that appropriate leveraging of history guidance provides an effective mechanism for motion planning.

Our contributions can be summarized as follows:

\begin{itemize}
\item[$\bullet$] Diffusion Forcing planning paradigm. We propose a chunk-level diffusion training scheme that jointly predicts history and future, and randomizes chunkwise timesteps. This forces the model to learn a reasonable causal dependency among history, future, and environment.
\item[$\bullet$] History-annealed CFG inference. We construct a history-annealed branch and a history-noise branch at inference, then fuse their outputs with a tunable coefficient to balance stability and flexibility.
\item[$\bullet$] Strong empirical results. Closed-loop evaluations and ablation studies on nuPlan demonstrate that DFP achieves performance competitive with state-of-the-art methods while producing stable and controllable planned trajectories in complex interactive driving scenarios.
\end{itemize}

%% file: sec/2_formatting.tex
\section{Related Work}

\paragraph{Imitation Learning and History-Induced Causal Confusion.}
Recent end-to-end imitation learning frameworks \cite{bansal2018chauffeurnet,kendall2019learning,chitta2022transfuser, jiang2023vad,chen2024vadv2,huang2023gameformer,hwang2024emma,chen2024end,sun2025generalizing,cheng2024pluto} have achieved remarkable progress by directly mapping sensory observations to control actions. Early milestones \cite{bansal2018chauffeurnet} demonstrate the feasibility of neural policies that learn from human demonstrations, while more recent works \cite{chen2024vadv2, cheng2024pluto, li2024hydra, hwang2024emma} have pushed performance through probabilistic planning and query-based decoding. However, conditioning on ego vehicle history including past positions, velocities, or trajectory segments, often induces causal confusion \cite{de2019causal, li2024ego,cheng2024rethinking}: the model learns to reproduce motion patterns rather than infer true causal relationships between scene context and actions. This leads to degraded closed-loop performance under distribution shift, as historical signals act as misleading signals \cite{li2024ego}. Methods like PlanTF \cite{cheng2024rethinking} incorporate history through careful architectural design and dropout mechanisms, other methods like \cite{diffusiondrive, xing2025goalflow, zheng2025diffusionbased, tan2025flow, li2024hydra} explicitly discard ego-history to avoid bias, sacrificing temporal coherence. We argue that ego history should not be discarded, but dynamically modulated via diffusion-based guidance to disentangle temporal consistency from real-time responsiveness.

\paragraph{Diffusion Models as Trajectory Generators.}
Diffusion models \cite{ho2020denoising} have revolutionized generative modeling for continuous domains, offering unique advantages for motion planning: native multi-modal uncertainty representations, flexible conditioning and guidance, and high sample quality. Recent work like Diffuser \cite{janner2022diffuser} and Diffusion Policy \cite{chi2025diffusion} introduce trajectory-level generation via denoising \cite{ajay2023is} and demonstrate remarkable success in visuomotor control.  
In autonomous driving, existing methods \cite{tan2025flow, zheng2025diffusionbased, diffusiondrive, xing2025goalflow} established strong baselines but omit ego vehicle history entirely. Critically, no existing method provides inference-time control over historical influence. Our work extends these foundations by introducing adaptive history modulation via block-wise noise scheduling and classifier-free guidance\cite{ho2022classifier}, preserving temporal consistency while enabling real-time responsiveness.

\paragraph{Temporal Consistency via Guided Diffusion Scheduling.}
Some methods enhance temporal continuity through structured generation or post-hoc regularization, e.g., real-time action chunking \cite{black2025real} employs soft masking to encourage alignment with past actions, Streaming Flow Policy \cite{jiang2025streaming} initializes ODE integration from recent actions without adjusting their influence during generation, while Past Token Prediction \cite{torne2025learninglongcontextdiffusionpolicies} and Bidirectional Decoding \cite{liu2024bid} select among sampled trajectories based on fixed criteria (e.g., past action consistency or future plan alignment). These methods treat temporal consistency as a correction layer applied after generation. They do not provide mechanisms to dynamically adjust the influence of historical signals based on environment context during generation.

\begin{figure*}[t]
\centering
\includegraphics[width=\linewidth]{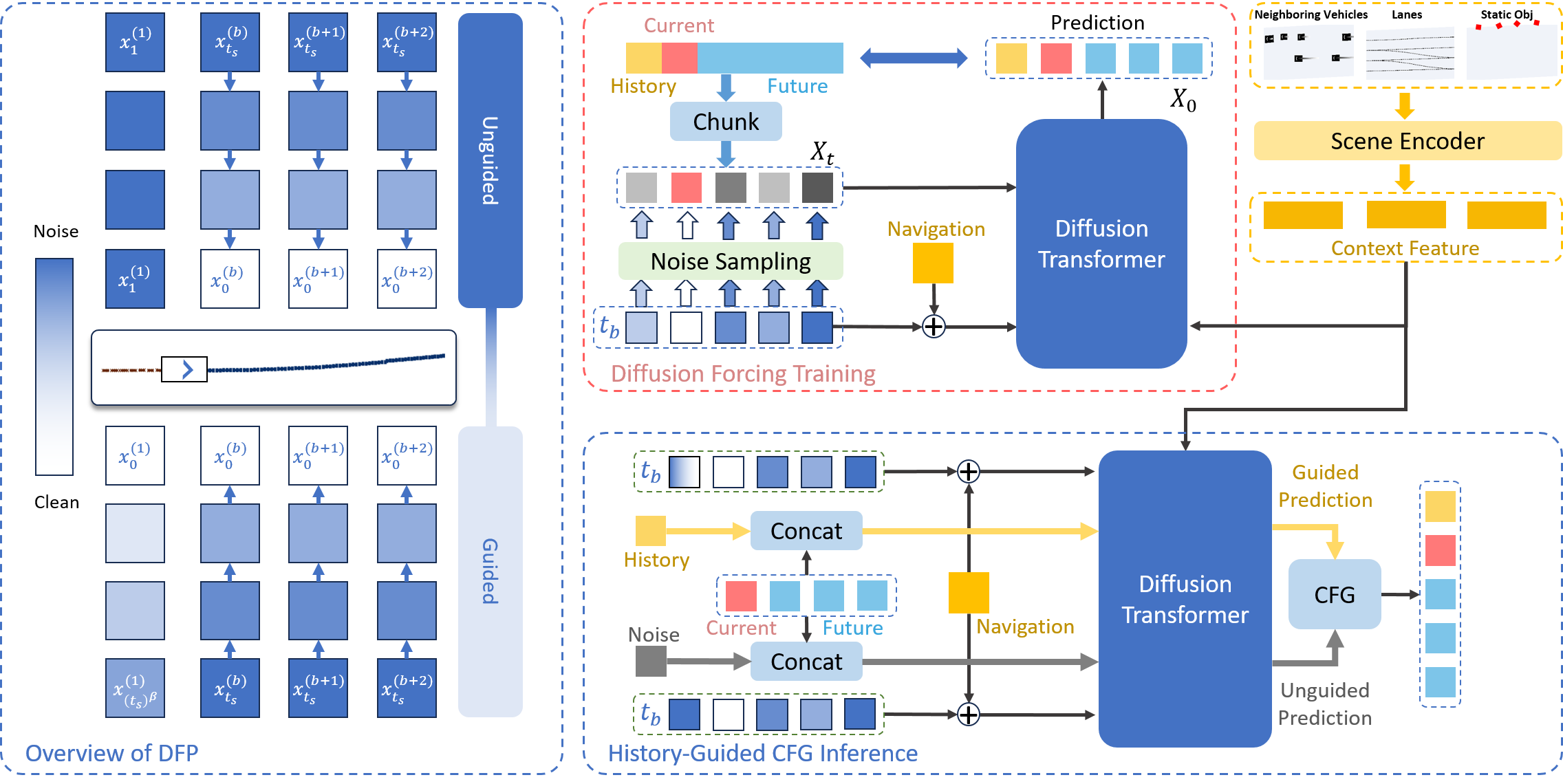}
\caption{\textbf{Overview of the \textit{Diffusion Forcing Planner} framework.}}
\label{fig:Overview}
\end{figure*}

Video generation research has pioneered integrated approaches. History-Guided Video Diffusion \cite{song2025history} uses noising-as-masking to anneal historical segments, achieving remarkable long-term coherence. Our key insight is that motion planning shares the same causal structure, but the history must be modulated in the presence of strong scene context, requiring controllable guidance strength. 
Unlike \cite{jiang2025streaming} which uses history only for initialization, or \cite{liu2024bid} which only selects among samples, we integrate history into the diffusion process itself through block-wise independent noise scheduling and classifier-free guidance, enabling online modulation of historical influence to adaptively trade off temporal coherence against real-time responsiveness.

%% file: sec/3_finalcopy.tex
\section{Method}

In this section, we briefly introduce the conditionally guided trajectory generation problem and present Diffusion Forcing Planner (DFP), a diffusion policy that leverages history-guided conditioning without over-fitting to the past. We formalize the task and then detail the two key components of DFP: (i) a Diffusion Forcing training scheme with block-wise noising and (ii) a history-annealed CFG inference procedure with annealed history. An overview of DFP is shown in Figure \ref{fig:Overview}.

\subsection{Problem Formulation}
We consider a trajectory prediction task in the autonomous driving setting. Given scene context ${C}$ — including surrounding agents, static objects, lane information and navigation — together with the history information ${H}$, the goal is to conditionally generate a future planning trajectory.

Formally, we aim to transform a source distribution $p_0(x_0)$ into a target distribution $q(x_1|C,H,w)$ along a probabilistic path. Here, $x_1$ denotes the generated future trajectory, and $w$ is a guidance factor that modulates how strongly the historical trajectory influences the generated plan. We interpret the generated future trajectory as the causal successor of $(C,H)$. Under this view, the learning objective is to capture a well-calibrated dependency among history, future, and environment while avoiding degenerate copying of $H$ into x.

\subsection{Training with Diffusion Forcing}
We define a \textbf{Chunk} as representing a trajectory $x_{1:S}$ with multiple tokens, where each token summarizes a contiguous subsequence of length $L$ ($1<L<S$). In order to operationalize the noising-as-masking strategy introduced in diffusion forcing, we partition the full trajectory into history and future chunks and assign independent noise levels to different chunks. Let the full trajectory be:
\begin{equation}
    x_0=[x^{-H}_0, ...,x^0_0,...,x^F_0]\in \mathbb{R}^{S\times 4}, \; S=H+1+F
\end{equation}
where each state is a 4-tuple including coordinates and heading. We uniformly split $x_0$ into $N$ consecutive chunks of length $L$. In practice, since the current state contains only a single timestep, we replicate it to the chunk length $L$ to enable chunkwise processing. Chunks are classified by the time span they cover: $N_H$ history chunks, one current chunk, and $N_F$ future chunks. By construction, no chunk mixes history and future points. Denote the index set of chunk $b$ by $\mathcal{I}_b$ and its clean sub-sequence by:
\begin{equation}
    x_0^{(b)}=[x^{(b)}_{0,i}]_{i \in \mathcal{I}_b}\in \mathbb{R}^{1\times 4L}, \; b=1,2,...,N
\end{equation}
where $i$ denotes the index of the trajectory point within each chunk.
For Diffusion Forcing, each block is perturbed with an independently sampled noise level $t_b \sim U(0,1)$ for each block and perturb it via the SDE marginal:

\begin{equation}
    x_{t_b}^{(b)}=\alpha(t_b)x_{0}^{(b)}+\sigma(t_b)\varepsilon^{(b)} , \; \varepsilon \sim \mathcal{N}(0,1)
\end{equation}
The model input is the concatenation $X_t=[x_{t_1}^{(1)} \:;\: x_{t_2}^{(2)}\:;\:...\:;\:x_{t_N}^{(N)}]$ together with per-chunk diffusion time $t=[t_1\;;\;t_2\;;\;...\;;\;t_N]$ and global conditions $(C,H)$. We train a diffusion transformer $f_\theta$ in the parameterization of $x_0$-prediction. Let $\hat{x}^{(b)}=f_\theta(X_t,t,C)_{|\mathcal{I}_b}$ denote the slice of the network output corresponding to block $b$. Before training, we set the noise level of the current block $t_{cur}=0$, yielding a hard boundary that anchors the plan. The diffusion-forcing loss sums chunk-wise objectives with weights $\lambda_{hist}$ and $\lambda_{futr}$:
\begin{equation}
\begin{split}
\mathcal{L}_{\text{denoise}}
= &\; \frac{\lambda_{\mathrm{hist}}}{N_H}\sum_{b = 1}^{N_H}
\mathbb{E}_{\,t_b,\, x^{(b)}_0}
\!\left[\bigl\| \widehat{x}^{(b)} - x^{(b)}_0 \bigr\|_2^2\right]
\\
&\;+\frac{\lambda_{\mathrm{futr}}}{N_F}\sum_{b = N_H+2}^{N}
\mathbb{E}_{\,t_b,\, x^{(b)}_0}
\!\left[\bigl\| \widehat{x}^{(b)} - x^{(b)}_0 \bigr\|_2^2\right]
\end{split}
\end{equation}
Randomizing ${t_b}$ implements noising-as-masking, in which large $t_b$
 masks a block in a strong noise, small $t_b$ exposes it clean. This forces the network to learn stable, causal conditioning under mixed availability of history and future information.

\subsection{History-Annealed CFG Inference}
We use classifier-free guidance (CFG) to modulate how strongly history affects the generated future. Inference proceeds with two branches, unguided and guided, that share the same sampler but differ in how the history chunk is formed. All future chunks are initialized from noise and concatenated with the clean current block to impose a hard boundary. At denoising step $s$ with global time $t_s\in[0,1]$, we assemble the per-chunk inputs $X_{t_s}=[x_{t_0}^{(0)}\:;\:x_{t_s}^{(1)}\:;\:...\:;\:x_{t_s}^{(N_F)}]$. For the current chunk, we set $t_0=0$ to kept it clean. The two branches differ on the history chunk. 

For unguided branch, we replace history chunk with pure noise $\varepsilon \sim \mathcal{N}(0,1)$ at every diffusion step to prevent any history signal leakage. Since the history block is pure noise, its associated time is reset to 1, while the time for future chunk decrease along the diffusion process. We set $t=\{1,...1, t_0, t_s,t_s,...,t_s\}$, which matches our training setup of chunkwise independent timesteps. In this branch, wo obtain the unguided prediction:
\begin{equation}
    \hat{X}_{0,unguided}=f_\theta([\varepsilon,X_{t_s}],t,C)
\end{equation}

For guided branch, the clean history $X_{history}$ is concatenated with $X_{t_s}$. To prevent the future from being over-constrained by history, we anneal the history with noise over diffusion steps. The history chunks start near noise and then rapidly return to the clean value. Concretely, the history chunks follow an annealing schedule:
\begin{equation}
    X_{guidance}=\alpha(t)X_{history}+\sigma(t)\varepsilon , \;t=(t_s)^\beta
    \label{denealed}
\end{equation}
 Here $(t_s)^\beta\:(\beta \ge 1)$ makes the early steps closer to noise and the final steps closer to the ground truth. The guidance chunks $X_{guidance}$ will be concatenated with $X_{t_s}$ at each step. In this branch, wo obtain the guided prediction:
 \begin{equation}
    \hat{X}_{0,guided}=f_\theta(X_{guidance}\;;\;X_{t_s},t|C)
\end{equation}

We apply CFG to fuse the predictions from the guided and unguided branches:
\begin{equation}
    \hat{X}_{0} = \hat{X}_{0,unguided} + w \bigl( \hat{X}_{0,guided} - \hat{X}_{0,unguided} \bigr)
    \label{CFG}
\end{equation}
where \(w \in [0,1]\) controls the influence of history guidance on the final prediction.

The final future trajectory is obtained by stitching the predicted future blocks back into a single sequence, with linear feathering applied on the overlap between adjacent blocks to ensure smooth transitions.

\subsection{Model Architecture}
We instantiate DFP as a chunk-wise diffusion transformer tailored to the ego-trajectory generation with history guidance. The network operates on a short sequence of trajectory tokens composed of history, current and future. 

Given normalized inputs $x \in \mathbb{R}^{B \times S \times (4L)}$, where $S$ denotes the number of tokens and $L$ denotes the number of trajectory points covered by each token. For a fair comparison, we adopt the same encoder architecture as Diffusion Planner to encode the scene context $C$ and the navigation route $R$.

\paragraph{Token-level positional embedding.} We treat each chunk in the trajectory as a token and assign it a learnable positional embedding that encodes its location within the sequence. The raw trajectory input $x$ is linearly projected to the model dimension $D$, and the resulting token features are added to a bank of learnable positional embeddings $x_{embedding} \in \mathbb{R}^{B \times S\times D}$:
\begin{equation}
    \tilde{x}=MLP(x)+x_{embedding}
\end{equation}

\paragraph{Per-token temporal embedding.} To support diffusion forcing with token-wise noise strengths, we equip each token with its own time embedding. We embed a per-token $t \in \mathbb{R}^{B \times S}$ through a sinusoidal Fourier feature mapping, and the resulting features are further transformed by a small MLP into a $D$-dimensional time embedding $t_{embedding}$. The navigation information $R$ is broadcast to each token and summed with 
$t_{embedding}$ to form the conditioning vector $y \in \mathbb{R}^{B \times S\times D}$.

\newcommand{\g}[1]{\textcolor{gray}{#1}}

\begin{table*}[ht]
\centering
\caption{\textbf{Main results on the nuPlan benchmarks.}
Overall score (\%) in non-reactive (NR) and reactive (R) closed-loop evaluations on \emph{Val14}, \emph{Test14}, and \emph{Test14-hard}. Raw model outputs are evaluated without additional post-processing for fair comparison. *: The best result from our multiple reimplementation attempts, slightly differing from the original report. \textbf{DFP-FM} combines DFP with the Flow Matching sampler.}
\begin{tabular*}{\textwidth}{@{\extracolsep{\fill}}llcccccc@{}}
\toprule
\multirow{2}{*}{\textbf{Type}} & \multirow{2}{*}{\textbf{Planner}}
& \multicolumn{2}{c}{\textbf{Val14}} & \multicolumn{2}{c}{\textbf{Test14}} & \multicolumn{2}{c}{\textbf{Test14-hard}} \\
\cmidrule(lr){3-4}\cmidrule(lr){5-6}\cmidrule(lr){7-8}
& & \textbf{NR} & \textbf{R} & \textbf{NR} & \textbf{R} & \textbf{NR} & \textbf{R} \\
\midrule
\g{Expert} & \g{Log-replay} & \g{93.53} & \g{80.32} & \g{85.96} & \g{68.80} & \g{94.03} & \g{75.86} \\
\midrule
\multirow{10}{*}{Learning-based}
& PDM-Open   & 53.53 & 54.24  & 52.81 & 57.23 & 33.51 & 35.83 \\
& UrbanDriver & 68.57 & 64.11 & 51.83 & 67.15 & 50.40 & 49.95 \\
& GameFormer & 13.32 & 8.69 & 11.36 & 9.31 & 7.08 & 6.69 \\
& PlanTF & 84.27 & 76.95 & 85.62 & 79.58 & 69.70 & 61.61 \\
& PLUTO & 88.89 & 78.11 & 89.90 & 78.62 & 70.03 & 59.74 \\
& CoPlanner & 89.48 & 79.00 & 90.31 & 78.81 & 76.82 & 64.47 \\
& Diffusion Planner & 89.87 & \textbf{82.80} & 89.19 & 82.93 & 75.99 & \textbf{69.22} \\
& Diffusion Planner\textsuperscript{*} & 87.87 & 77.48 & 90.01 & 79.61 & 74.26 & 61.25 \\
\midrule
& \textbf{DFP(ours)} & 90.33 & 79.97 & \textbf{90.69} & 81.96 & 76.91 & 63.56 \\
& \textbf{DFP-FM(ours)} & \textbf{92.68} & 81.30 & 90.62 & \textbf{83.59} & \textbf{79.43} & 67.94 \\
\bottomrule
\end{tabular*}
\label{tab:main_results}
\end{table*}

\paragraph{DiT Blocks.} The decoder backbone stacks multiple DiT blocks. Each block applies adaptive LayerNorm(adaLN) FiLM-style conditioning from $y$ to both self-attention and MLP residuals, and a cross-attention to the fused perception context $C$. We apply Multi-Head Self-Attention(MHSA) operates along the token axis—spanning history, current, and future chunks—to capture long-range dependencies and enable information flow and mutual regularization across chunks:
\begin{equation}
    \tilde{x}=\tilde{x}+MHSA(adaLN(\tilde{x},y))
\end{equation}
Multi-Head Cross-Attention (MHCA) attends from per-token queries to the contextual memory $C$ to injecting scene priors into every token:
\begin{equation}
    \tilde{x}=\tilde{x}+MHCA(adaLN(\tilde{x},y),C)
\end{equation}

Finally, an adaLN-modulated projection maps each token back to the corresponding trajectory chunk.

%% file: sec/4_Experiments.tex
\section{Experiments}

\subsection{Experimental Setup}
\paragraph{Benchmarks.} We evaluate DFP on nuPlan, a large-scale, real-world dataset and closed-loop simulation suite purpose-built for autonomous driving planning. NuPlan provides over 1,200 hours of expert driving across diverse road types, traffic densities, and multi-agent interactions, enabling rigorous assessment of long-horizon, interaction-aware planning. Following the official nuPlan protocol, we report results in both non-reactive (log-replay of other agents) and reactive (agents controlled by IDM policy\cite{kesting2013traffic}) simulation settings. We adopt three standard nuPlan splits: (1) Val14. A validation set of 1,118 scenarios used for model selection and ablations. (2) Test14-random. A test set of over 200 scenarios randomly sampled from the nuPlan Planning Challenge, providing unbiased performance estimates across scene types. (3) A curated test set of 272 challenging scenarios identified as worst-cases under the rule-based PDM baseline, stressing rare and safety-critical interactions. All evaluations are closed-loop and use the official simulator configuration. We argue that this tri-partite benchmark offers a comprehensive and fair measure of a planner’s stability, adaptivity, and interaction competence.

\begin{table*}[htbp]
\centering
\caption{\textbf{Case study on scenario-specific metrics.} Evaluation is conducted on the nuPlan \texttt{Val14} benchmark in the non-reactive setting. The number following each scenario type indicates the count of samples belonging to that scene.
``DP'' denotes the \textbf{Diffusion Planner} baseline; \textbf{Score} denotes the overall score for the scenario; \textbf{Collision}/\textbf{TTC}/\textbf{Drivable}/\textbf{Comfort}/\textbf{Progress} denote the respective per-scenario metrics. All per-scenario metrics are Boolean indicators, reported as the fraction of cases satisfying the condition (higher is better).}
\begin{tabular*}{\textwidth}{@{\extracolsep{\fill}}l c c c c c c c@{}}
\toprule
\textbf{Scenario type} & \textbf{Planner} & \textbf{Score} & \textbf{Collision} & \textbf{TTC} & \textbf{Drivable} & \textbf{Comfort} & \textbf{Progress} \\
\midrule

\multirow{3}{*}{All(1118)}
  & PlanTF              & 84.27 & 94.01 & 89.00 & 95.89 & 93.02 & 98.75 \\
  & DP                  & 87.80 & 95.53 & \textbf{92.13} & \textbf{97.32} & 91.86 & \textbf{100.00} \\
  & \textbf{DFP (ours)} & \textbf{90.33} & \textbf{96.60} & 91.86 & 92.49 & \textbf{96.69} & 99.91 \\
\midrule

\multirow{3}{*}{Low magnitude speed(100)}
  & PlanTF              & 85.00 & 96.00 & 92.00 & 95.00 & 89.00 & 95.00 \\
  & DP                  & 86.51 & 97.00 & 97.00 & 95.00 & 94.00 & 96.00 \\
  & \textbf{DFP (ours)} & \textbf{91.08} & \textbf{98.00} & \textbf{97.00} & \textbf{97.00} & \textbf{96.00} & \textbf{99.00} \\[0.5em]

\multirow{3}{*}{High magnitude speed(99)}
  & PlanTF              & 89.39 & 94.95 & 94.95 & 95.96 & 94.95 & 98.99 \\
  & DP                  & 84.50 & 95.96 & 95.96 & 96.97 & 60.61 & 100.00 \\
  & \textbf{DFP (ours)} & \textbf{94.95} & \textbf{98.99} & \textbf{97.98} & \textbf{97.98} & \textbf{96.97} & \textbf{100.00} \\
\midrule

\multirow{3}{*}{Starting left turn(100)}
  & PlanTF              & 80.42 & 93.00 & 85.00 & 98.00 & 81.00 & 98.00 \\
  & DP                  & 82.38 & 93.00 & \textbf{90.00} & 97.00 & 90.00 & 99.00 \\
  & \textbf{DFP (ours)} & \textbf{86.74} & \textbf{93.00} & 88.00 & \textbf{98.00} & \textbf{94.00} & \textbf{100.00} \\[0.5em]

\multirow{3}{*}{Starting right turn(98)}
  & PlanTF              & 70.13 & 91.84 & 81.63 & 90.82 & 91.84 & 94.90 \\
  & DP                  & 78.16 & 89.80 & 81.63 & \textbf{94.90} & 94.90 & 100.00 \\
  & \textbf{DFP (ours)} & \textbf{79.38} & \textbf{94.90} & \textbf{82.65} & 88.78 & \textbf{95.92} & \textbf{100.00} \\
\bottomrule
\end{tabular*}
\label{tab:case_study}
\end{table*}

\paragraph{Baseline.} We compare DFP with several representative learning-based planners. For all methods, we directly use the raw model outputs as the final trajectories for evaluation, without any additional post-processing modules. 
\begin{itemize}
\item[$\bullet$] \textbf{PDM-Open\cite{dauner2023parting}.} A straightforward MLP that predicts future waypoints, corresponding to the learning-based version of PDM, the first-place winner of the nuPlan challenge.
\item[$\bullet$] \textbf{UrbanDriver\cite{scheel2022urban}.} A learning-based planner implemented in nuPlan that optimizes a driving policy via policy-gradient methods on closed-loop rollouts.
\item[$\bullet$] \textbf{GameFormer\cite{huang2023gameformer}.} A game-theoretic transformer that models interactions between the ego vehicle and surrounding agents as a multi-agent game for interactive planning.
\item[$\bullet$] \textbf{PlanTF\cite{cheng2024rethinking}.} A simple and purely imitation-based planner built on a transformer architecture, which explores lightweight designs tailored to closed-loop motion planning.
\item[$\bullet$] \textbf{PLUTO\cite{cheng2024pluto}.} A query-based model that jointly handles lateral and longitudinal maneuvers, enabling flexible and diverse driving behaviors within a unified planning architecture.
\item[$\bullet$] \textbf{CoPlanner\cite{zhong2025coplanner}.} A unified framework that simultaneously models multi-agent interactive trajectory generation and contingency-aware motion planning for the ego vehicle.
\item[$\bullet$] \textbf{Diffusion Planner\cite{zheng2025diffusionbased}.} A diffusion-based planner that fully exploits diffusion models with a dedicated architecture for high-performance motion planning. DFP is built on top of Diffusion Planner(referred to as \textit{DP} in the following text).
\end{itemize}

\subsection{Implementation Details}
We train DFP on 1M nuPlan clips. Each clip covers 2s of history and 8s of future, sampled at 10 Hz. Each trajectory point consists of position and heading, encoded as $(x,y,cos\theta,sin\theta)$. 

Following Diffusion Planner, we express all geometry in an ego-centric frame whose origin is the ego-current pose with heading aligned to the x-axis. We apply per-channel z-score normalization using means and standard deviations estimated on the training set and fixed for both training and inference. Besides, a lightweight, geometry-consistent augmentation is adopted by perturbing the current state with applying interpolation to both history and future trajectories. 

During training, we sample the noise level $t$ for the history segment from a Beta distribution, so that more samples concentrate near the two extremes $t\approx0$ (clean) and $t\approx1$ (pure noise). This encourages the model to frequently see the near-clean and near-noise regimes of history during training, aligning with the configurations used at inference time. During chunking, each full trajectory is divided into 
$N=6$ chunks, with $L=20$ points per chunk (the current chunk is formed by repeating the single current state to 20 points). All models are trained in the $x_0$-prediction parameterization. At inference, we adopt DPM-Solver\cite{lu2022dpm} for fast sampling. DFP is trained with a global batch size of 2048 for 500 epochs, using a 5-epoch warmup phase. We use the AdamW optimizer with a learning rate of $2e-4$.

\begin{table*}[ht]
\centering
\caption{\textbf{Effect of designs in the diffusion decoder.} 
Ablations on \texttt{Val14} under Non-Reactive (NR) and Reactive (R) evaluation. }
\label{tab:ablation_diffusion_decoder}
\begin{tabular*}{\textwidth}{@{\extracolsep{\fill}} l|cccc cc @{}} 
\toprule
\multirow{2}{*}{\textbf{ID}} 
  & \multicolumn{4}{c}{\textbf{Component}} 
  & \multicolumn{2}{c}{\textbf{Val14 Score}} 
   \\
\cmidrule(lr){2-5}\cmidrule(lr){6-7}
  & \textbf{Diffusion Forcing} 
  & \textbf{Chunk} 
  & \textbf{History Guidance} 
  & \textbf{Annealed History} 
  & \textbf{NR} 
  & \textbf{R} 
   \\ 
\midrule
A1 & \xmark       & \xmark       & \xmark       & \xmark       & 87.87 & 77.48  \\
A2 & \cmark   & \xmark       &  \xmark       & \xmark       & 85.18 & 75.27   \\
A3 & \xmark   & \cmark       &  \xmark       & \xmark       & 87.58 & 77.10   \\
A4 & \cmark   &\cmark   & \xmark       & \xmark       & 88.79 & 77.49  \\
A5 & \cmark   & \xmark   & \cmark   & \xmark       & 86.56 & 75.93  \\
A6 & \cmark   & \cmark  & \cmark   & \xmark   & 89.24 & 79.16  \\
A7 & \cmark 
   & \cmark 
   & \cmark 
   & \cmark 
   & \textbf{90.33} & \textbf{79.97}  \\
\bottomrule
\end{tabular*}
\end{table*}

\begin{figure*}[t]
\centering
\includegraphics[width=\linewidth]{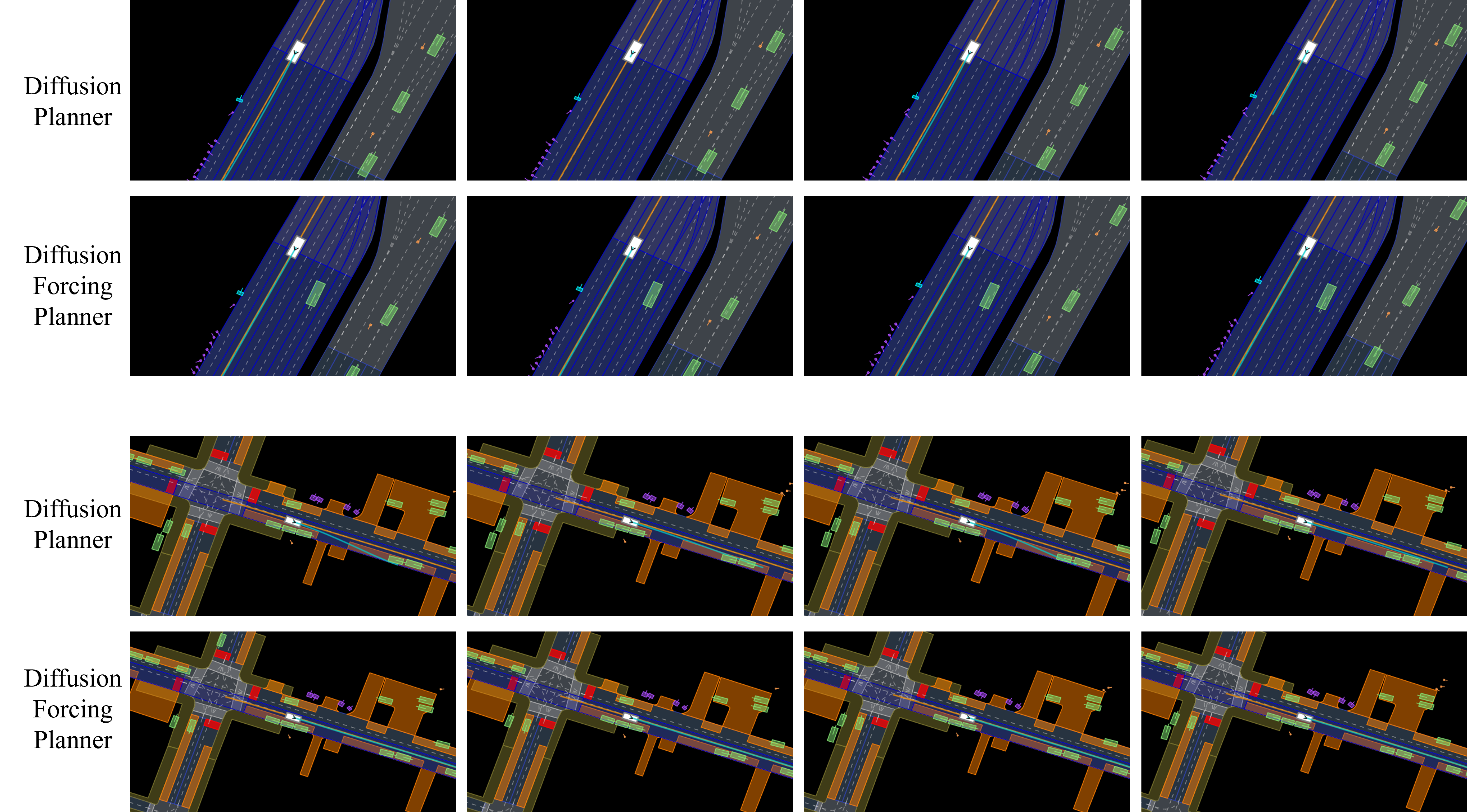}
\caption{\textbf{DP vs.\ DFP qualitative comparison.} The figure visualizes trajectory predictions over four consecutive frames in two scenarios. Yellow trajectories denote expert (log-replay) trajectories and blue trajectories denote model predictions. Compared to DP, DFP maintains smoother and more temporally consistent trajectories across frames.}
\label{fig:vis}
\end{figure*}

\subsection{Main results and Case Study}

\paragraph{Main Results.} Table \ref{tab:main_results} reports a comprehensive comparison between DFP and other learning-based baselines on the nuPlan benchmarks. We additionally evaluated DFP combined with the Flow Matching sampler (denoted as \textbf{DFP-FM} in Table \ref{tab:main_results}), with details provided in the supplementary material. To ensure a fair evaluation, all methods are evaluated without any additional post-processing. We directly use the raw model outputs as the final trajectories for scoring. Across multiple nuPlan benchmarks, DFP surpasses previous state-of-the-art approaches. In particular, compared with UrbanDriver, GameFormer, and PlanTF, DFP achieves substantial improvements on most metrics, demonstrating strong planning capability. Moreover, starting from Diffusion Planner as a diffusion-based baseline, DFP consistently yields further gains, validating that our design more effectively exploits diffusion models for motion planning. 

Concretely, on \emph{Val14} DFP improves the non-reactive score from $87.87$ (Diffusion Planner) to $90.33$ and the reactive score from $77.48$ to $79.97$ (\,$+2.46$ and $+2.49$ points, respectively).
On \emph{Test14} DFP improves over Diffusion Planner by $+2.35$ points on NR challenge, reaching 80+ performance. 
On the more challenging \emph{Test14-hard} benchmark, DFP attains $76.91$ (NR), exceeding Diffusion Planner by $+2.65$ points and slightly surpassing CoPlanner.



\paragraph{Case Study.} Table \ref{tab:case_study} presents a case study comparing DFP and the baseline method across several specific driving scenarios and evaluation metrics. The \textbf{All} row at the top of table reports the aggregate performance of all 1118 scenarios in nuPlan Val14 benchmark. This provides a holistic view of closed-loop planning quality across diverse traffic conditions.
The middle of the table lists two state-preserving scenario types, which are used to assess how stable and consistent the generated trajectories remain when the driving state should be maintained. The bottom of the table lists two environment-changing scenario types, which test whether the model can adapt its strategy when the environment or interaction pattern changes. As shown in Table \ref{tab:case_study}, DFP achieves consistent improvements over Diffusion Planner across all scenario types, with varying degrees of gain depending on the scenario.

Specifically, in stable scenarios, DFP improves the overall score by \textbf{+4.57 points} on \textit{low magnitude speed} and \textbf{+10.45 points} on \textit{high magnitude speed}. Notably, in the high-speed setting, DFP attains a \textit{Comfort} score of \textbf{96.97}, exceeding DP by \textbf{30+ points}, indicating substantially smoother, less jittery trajectories under steady dynamics. Figure \ref{fig:vis} visualizes a nearly straight, constant-velocity case: DP exhibits inter-frame fluctuations, while DFP maintains stable heading and speed due to history-guided CFG. In turning scenes (\textit{starting left/right turn}), DFP also outperforms DP, benefiting from controllable history guidance that stabilizes motion while remaining responsive to evolving environments and interactions.

The results indicate that DFP not only stabilizes trajectory outputs in state-preserving regimes, but also produces more appropriate and adaptive plans under changing environments.

\subsection{Ablation Study}
\paragraph{Effect of designs in the diffusion decoder.} Table \ref{tab:ablation_diffusion_decoder} summarizes the effect of our design choices in the diffusion decoder. \textbf{A1} corresponds to the Diffusion Planner baseline, on top of which all our subsequent designs are incrementally built. 

\textbf{A2} adds only the Diffusion Forcing mechanism. In this initial design, we assign an independent noise level to every trajectory point. Each token along the trajectory receives its own diffusion timestep. With a chunk length of $L=1$, the entire horizon is represented by singleton chunks, each covering only one timestep, which leads to a drop in performance. \textbf{A3} demonstrates this: setting $L>1$ groups consecutive steps into chunks, and performance is restored. We attribute this result to the following factors: When every token corresponds to a single timestep ($L=1$), each token carries little trajectory semantics. Therefore the planner must stitch many noisy micro-steps into a coherent plan through long-range dependencies, making credit assignment harder.

\textbf{A4} integrates the settings of A2 and A3. Instead of assigning noise at the point level, we sample one noise level per chunk. This preserves the benefits of action chunking while simplifying the decoder structure and improving training stability. 

\textbf{A5} introduces history guidance without action chunking. In this setting, the gains over the baseline are modest, suggesting that simply injecting history guidance into a non-chunked decoder is insufficient to fully exploit historical information. 

\textbf{A6} combines action chunking with clean history guidance and yields gains in planning quality, indicating that history is beneficial when coupled with a chunk-based diffusion process. However, when the history guidance is kept persistently clean, it can become overly dominant in some scenarios. The policy lingers on historical motion patterns and reacts slowly.

To remedy this, \textbf{A7} additionally introduces the Annealed History mechanism. This variant leverages the model’s ability to predict history while injecting gradually annealed ground-truth history as guidance along the diffusion process. It balances the influence of history and allows the model to adapt its future plan when the scene changes.

\begin{figure}[t]
\centering
\includegraphics[width=\linewidth]{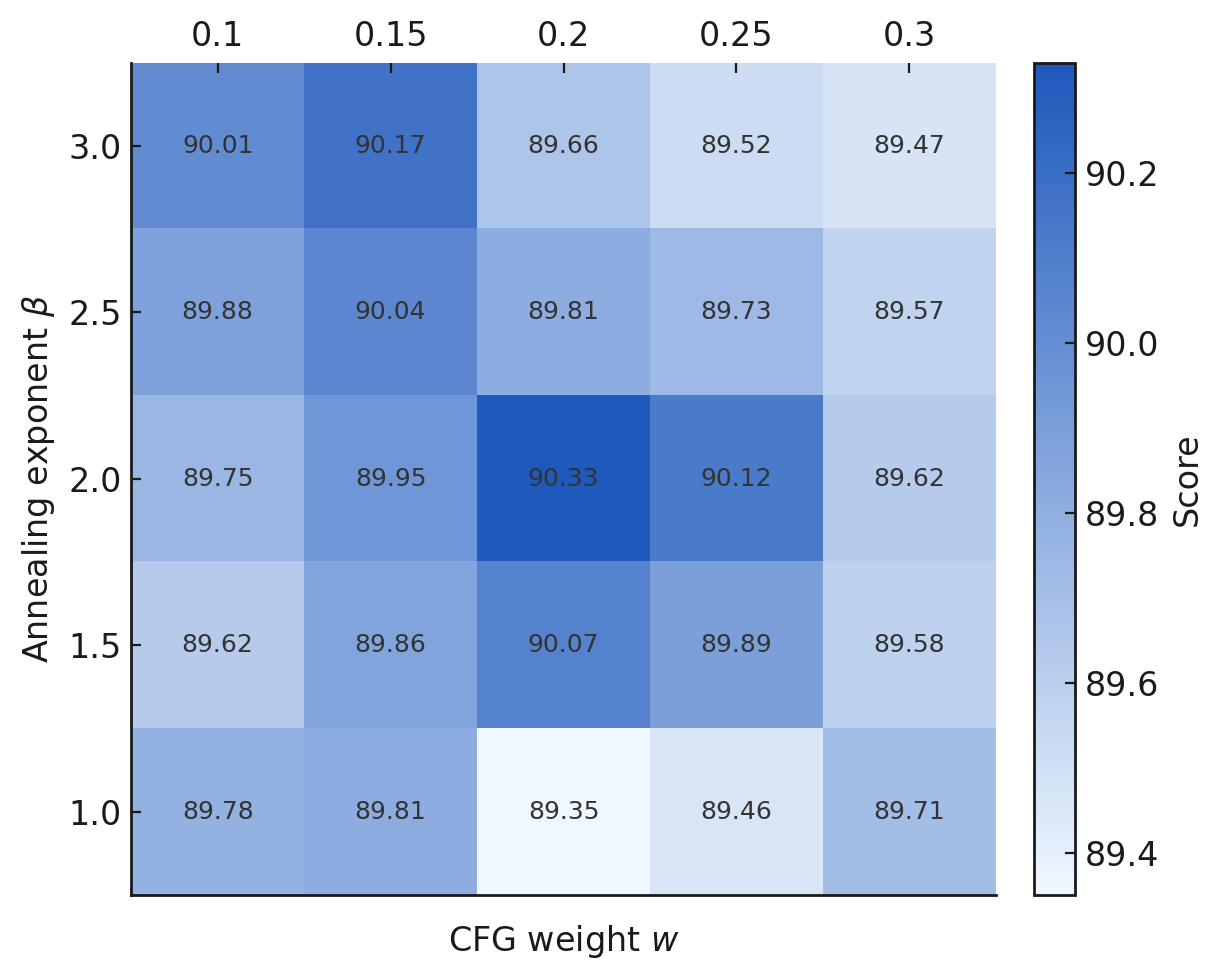}
\caption{\textbf{Effect of history guidance weights.}}
\label{fig:heatmap}
\end{figure}

\paragraph{History guidance weights.} Figure \ref{fig:heatmap} analyzes the sensitivity of DFP to the two hyperparameters used in history guidance. The horizontal axis $w$ denotes the strength of the history-guided CFG term (see Eq.\ref{CFG}, while the vertical axis $\beta$ controls the annealing speed of the ground-truth history schedule $(t_s)^{\beta}$ (see Eq.\ref{denealed}. The results show that both too weak and too strong history guidance hurt performance. The best trade-off is achieved at $w=0.2$ and $\beta=2.0$, where DFP attains the highest overall score.

%% file: sec/conclusion.tex
\section{Conclusion}
In this work, we introduced Diffusion Forcing Planner (DFP), a diffusion‐based planning policy that explicitly leverages history guidance while preserving controllability and robustness. We separately detail two core modules of DFP: diffusion-forcing training and history-guided CFG inference. During training, DFP learns causally consistent generation via chunked noising-as-masking. At inference, DFP anneals historical guidance and fuses guided and unguided branches to controllably influence future trajectories at test time. The results demonstrate DFP markedly improves comfort and inter-frame consistency, while retaining competitive closed-loop performance on nuPlan without relying on heavy post-processing. Despite these gains, DFP has limitations. Performance can be sensitive to the guidance hyperparameters. Future work will focus on applying adaptive guidance scheduling to end-to-end autonomous driving framework, and planning evaluations on broader benchmarks to assess generalization.

%% file: sec/X_suppl.tex
\begin{center}
    {\Large\bfseries Appendix}
\end{center}
\vspace{0.5em}


\section{Supplementary Methods}
\begin{figure}[htbp]
\centering
\includegraphics[width=\linewidth]{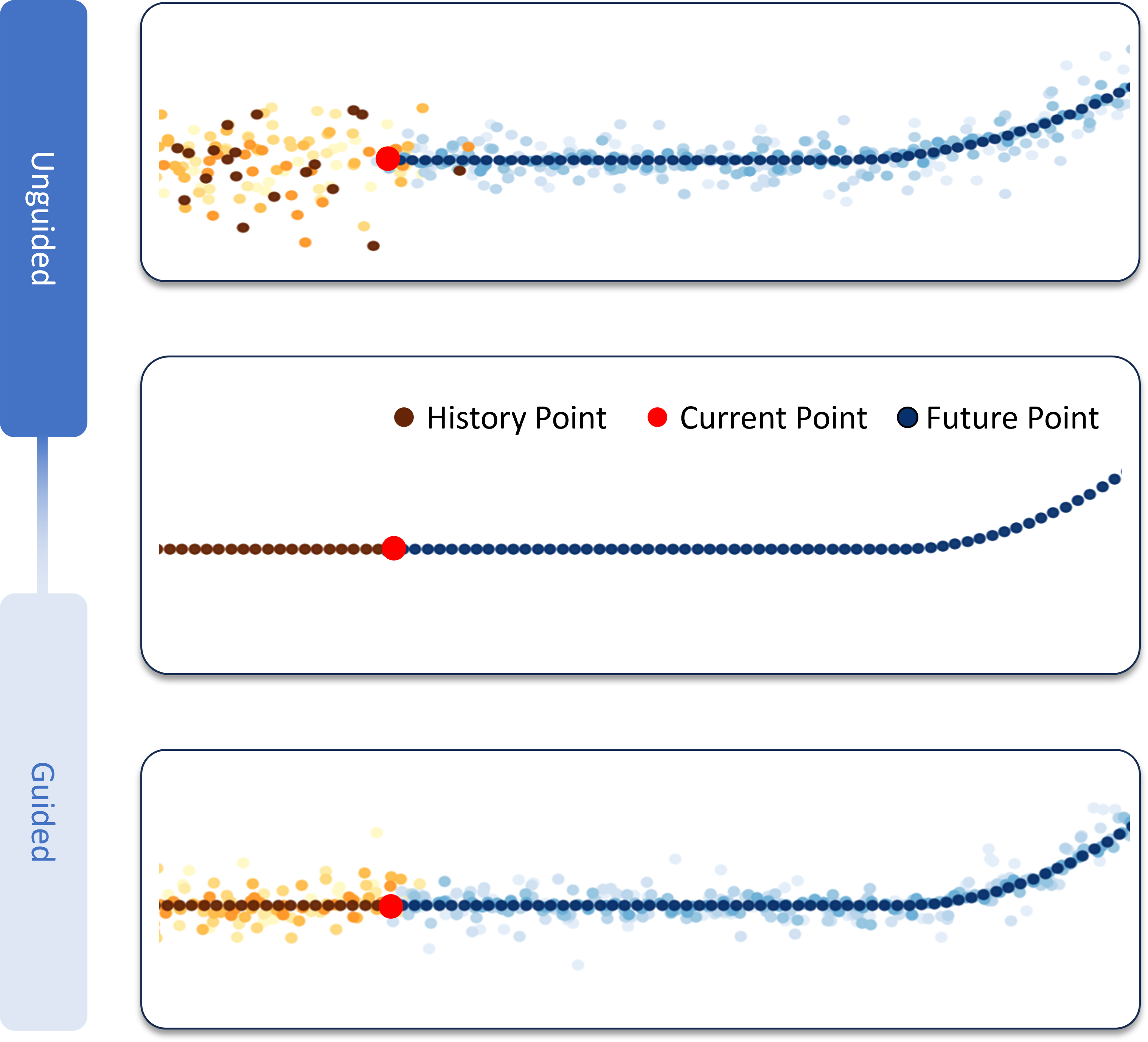}
\caption{\textbf{Annealed history CFG.} Lighter points indicate earlier diffusion timesteps.}
\label{fig:annealed}
\end{figure}
Algorithms 1 and 2 are presented in the Method section as pseudocode, and together they provide the complete pipelines of the two components of DFP: Algorithm 1 — Training with Diffusion Forcing, and Algorithm 2 — History-annealed CFG Inference. Figure\ref{fig:annealed} visualizes the annealed history CFG procedure.
\begin{algorithm}[htbp]
\caption{Diffusion-Forcing Training with Chunkwise Noising-as-Masking}
\label{alg:dfp-training}
\begin{algorithmic}[1]
\Require scene contexts $C$; GT trajectory $x_0=[x^{-H}_0,\dots,x^{0}_0,\dots,x^{F}_0]$, total length $S=H+1+F$; chunk length $L$, the number of chunks $N$ ($N_H$ history, $1$ current, $N_F$ future); SDE marginals $(\alpha_t,\sigma_t)$; DiT $f_\theta$; weights $\lambda_{\text{hist}},\lambda_{\text{futr}}$
\Ensure parameter update of $\theta$
\State Split $x_0$ into $N$ index sets $\{\mathcal{I}_b\}_{b=1}^{N}$, then each chunk $x^{(b)}_0=\{x_{0,i}\}_{i\in \mathcal{I}_b}$
\State Sample per-block noise levels $t_b$, $b=1,\dots,N$; set current chunk $t_{0}\leftarrow0$

\For{$b=1$ to $N$}
  \State Sample $\varepsilon^{(b)}\sim\mathcal{N}(0,I)$
  \State Noised chunk: $x^{(b)}_{t_b} \leftarrow \alpha_{t_b}\,x^{(b)}_0 + \sigma_{t_b}\,\varepsilon^{(b)}$
\EndFor
\State Pack tokens $X_t=\mathrm{concat}\big(\{x^{(b)}_{t_b}\}_{b=1}^{N}\big)$,\quad $t=\{t_b\}_{b=1}^{N}$
\State Predict per-block $\{\hat{x}^{(b)}_{t_b}\}_{b=1}^{N} \leftarrow f_\theta\!\big(X_t,\,t,\,C\big)$
\State \textbf{Reconstruction losses:}
\State $L_{\text{hist}} \leftarrow \frac{1}{N_H }\sum_{b=1}^{N_H}\sum_{i\in I_b}\!\big\lVert \hat{x}^{(b)}_{0,i}-x^{(b)}_{0,i}\big\rVert_2^2$
\State $L_{\text{futr}} \leftarrow \frac{1}{N_F }\sum_{b=N_H+2}^{N}\sum_{i\in I_b}\!\big\lVert \hat{x}^{(b)}_{0,i}-x^{(b)}_{0,i}\big\rVert_2^2$

\State Total loss: $ L \leftarrow \lambda_{\text{hist}}L_{\text{hist}}+\lambda_{\text{futr}}L_{\text{futr}}$
\State Update $\theta\leftarrow\theta-\eta\,\nabla_\theta L$
\end{algorithmic}
\end{algorithm}

\begin{algorithm}[htbp]
\caption{Annealed-History Classifier-Free Guidance (CFG) Inference}
\label{alg:dfp-infer}
\begin{algorithmic}[1]
\Require scene context $C$; clean history chunks $X_{\mathrm{history}}$; current chunk $x^{(0)}_{0}$; full length $S{=}H{+}1{+}F$; chunk length $L$; the number of chunks $N{=}N_H{+}1{+}N_F$; diffusion time $t_s$; guidance weight $w$; history anneal exponent $\beta$; SDE $(\alpha_t,\sigma_t)$; DiT $f_\theta$
\Ensure predicted future trajectory $\widehat{x}_{\mathrm{future}} {=} [x^{1}_0, x^{2}_0, \dots, x^{F}_0]$
\State Initialize future chunks with noise: $x^{(b)}_{t_s} \sim \mathcal{N}(0, I)$, for $b{=}1,2,\dots,N_F$
\State Form $X_{t_s} \gets \big[x^{(0)}_{0},\, x^{(1)}_{t_s},\, \dots,\, x^{(N_F)}_{t_s}\big]$
\For{$t = t_s,\, \dots,\, 0$}
  \Statex \textbf{Unguided branch}
  \State Sample history noise chunks $\varepsilon \sim \mathcal{N}(0,I)$
  \State $X \gets \big[\varepsilon, \, X_{t_s}\big]$
  \State $\mathbf{t} \gets \big[\,\underbrace{1,\dots,1}_{N_H},\, 0,\, \underbrace{t_s,\dots,t_s}_{N_F}\big]$
  \State $\widehat{X}_{0, unguided} \gets f_\theta\!\big(X,\, \mathbf{t},\, C\big)$
  \Statex \textbf{Guided branch}
  \State Sample $\varepsilon \sim \mathcal{N}(0,I)$;\; $t_{\mathrm{hist}} \gets (t_s)^{\beta}$
  \State $X_{\mathrm{guidance}} \gets \alpha_{t_{\mathrm{hist}}}\, X_{\mathrm{history}} + \sigma_{t_{\mathrm{hist}}}\, \varepsilon$
  \State $X \gets \big[X_{\mathrm{guidance}},\, X_{t_s}\big]$
  \State $\mathbf{t} \gets \big[\,\underbrace{t_{\mathrm{hist}},\dots,t_{\mathrm{hist}}}_{N_H},\, 0,\, \underbrace{t_s,\dots,t_s}_{N_F}\big]$
  \State $\widehat{X}_{0,guided} \gets f_\theta\!\big(X,\, \mathbf{t},\, C\big)$
  \Statex \textbf{CFG fusion}
  \State $\widehat{X}_{0} \gets \widehat{X}_{0,unguided} + w\big(\widehat{X}_{0,guided} - \widehat{X}_{0,unguided}\big)$
  \Statex \textbf{Hard constraint \& step update}
  \State Reset current chunk to $x^{(0)}_{0}$
\EndFor
\State Concatenate all future chunks to compose $\widehat{x}_{\mathrm{future}}$
\end{algorithmic}
\end{algorithm}

\section{Additional Experiments on Flow Matching and Mix-of-Transformer}
While our main approach utilizes a Diffusion-based decoder to ensure stability and history controllability, we recognize the potential of the Flow Matching paradigm for achieving straighter generation trajectories. In this exploratory study, we substituted the Diffusion with the Flow Matching \cite{tan2025flow}. Additionally, to enhance the model's capacity in capturing diverse multi-modal interactions, we upgrade our backbone to a Mixture-of-Transformer structure\cite{tan2025flow,black2024pi_0,liang2024mixture}. This design leverages multiple expert transformers to specialize in varied traffic contexts. As shown in Table \ref{tab:main_results_sup} and Table \ref{tab:case_study_sup}, our new baseline outperforms the original DFP framework, and achieves state-of-the-art closed-loop performance on nuPlan non-reactive benchmarks.
\begin{table*}[htbp]
\centering
\caption{\textbf{Case study on scenario-specific metrics.} Evaluation is conducted on the nuPlan \texttt{Val14} benchmark in the non-reactive setting.
*: We reproduce the baselines\cite{zheng2025diffusionbased} on the nuPlan\cite{nuplan}, and may differ slightly from the original report, likely due to differences in the randomly sampled 1M training subset. $\dagger$: Motivated by the design of Flow Planner \cite{tan2025flow}, we incorporate Flow Matching and a Mixture-of-Transformer architecture into our Diffusion Forcing Planner; \textbf{Score} denotes the overall score for the scenario; \textbf{Collision}/\textbf{TTC}/\textbf{Drivable}/\textbf{Comfort}/\textbf{Progress} denote the respective per-scenario metrics. All per-scenario metrics are Boolean indicators, reported as the fraction of cases satisfying the condition (higher is better).}
\begin{tabular*}{\textwidth}{@{\extracolsep{\fill}}l c c c c c c c@{}}
\toprule
\textbf{Scenario type} & \textbf{Planner} & \textbf{Score} & \textbf{Collision} & \textbf{TTC} & \textbf{Drivable} & \textbf{Comfort} & \textbf{Progress} \\
\midrule

\multirow{4}{*}{All}
& PlanTF              & 84.27 & 94.01 & 89.00 & 95.89 & 93.02 & 98.75 \\
  & Diffusion Planner*              & 87.80 & 95.53 & 92.13 &97.32 &91.86 & \textbf{100.00} \\
  & \textbf{DFP (ours)} & {90.33} & {96.60} & {91.86} & {92.49} & {96.69} & {99.91} \\
  & \textbf{DFP (ours) $\dagger$} & \textbf{92.68} & \textbf{98.03} & \textbf{93.20} & \textbf{98.12} & \textbf{98.48} & {99.82} \\
\midrule
\multirow{4}{*}{Low magnitude speed}
& PlanTF              & 85.00 & 96.00 & 92.00 & 95.00 & 89.00 & 95.00 \\
  & Diffusion Planner*              & 86.51 & 97.00 & 97.00 & 95.00 &94.00 & 96.00 \\
  & \textbf{DFP (ours)} & \textbf{91.08} & {98.00} & {97.00} & {97.00} & {96.00} & \textbf{99.00}  \\
  & \textbf{DFP (ours) $\dagger$} & {90.74} & \textbf{98.00} & \textbf{97.00} & \textbf{97.00} & \textbf{100.00} & 98.00 \\ [0.5em]
\multirow{4}{*}{High magnitude speed}
& PlanTF              & 89.39 & 94.95 & 94.95 & 95.96 & 94.95 & 98.99 \\
  & Diffusion Planner*              & 84.50 & 95.96 & 95.96 & 96.97 & 60.61 & 100.00 \\
  & \textbf{DFP (ours)} & {94.95} & {98.99} & {97.98} & {97.98} & {96.97} & {100.00} \\
  & \textbf{DFP (ours) $\dagger$} & \textbf{97.40} & \textbf{100.0} & \textbf{98.99} & \textbf{98.99} & \textbf{98.99} & \textbf{100.00} \\
\midrule
\multirow{4}{*}{Starting left turn}
& PlanTF              & 80.42 & 93.00 & 85.00 & 98.00 & 81.00 & 98.00 \\
  & Diffusion Planner*              & 82.38 & 93.00 & 90.00 & 97.00 & 90.00 & 99.00 \\
  & \textbf{DFP (ours)} & {86.74} & {93.00} & {88.00} & {98.00} & {94.00} & {100.00} \\
  & \textbf{DFP (ours) $\dagger$} & \textbf{93.15} & \textbf{97.00} & \textbf{93.00} & \textbf{100.00} & \textbf{99.00} & \textbf{100.00} \\[0.5em]
\multirow{3}{*}{Starting right turn}
& PlanTF              & 70.13 & 91.84 & 81.63 & 90.82 & 91.84 & 94.90 \\
  & Diffusion Planner*              & 78.16 & 89.80 & 81.63 & 94.90 & 94.90 & 100.00 \\
  & \textbf{DFP (ours)} & {79.38} & {94.90} & {82.65} & {88.78} & {95.92} & {100.00} \\
  & \textbf{DFP (ours) $\dagger$} & \textbf{86.61} & \textbf{96.94} & \textbf{83.67} & \textbf{95.92} & \textbf{97.96} & \textbf{100.00}  \\
\bottomrule
\end{tabular*}
\label{tab:case_study_sup}
\end{table*}

\newcommand{\gtext}[1]{\textcolor{gray}{#1}}
\begin{table*}[htbp]
\centering
\caption{\textbf{Main results on the nuPlan benchmarks.}
Overall score (\%) in non-reactive (NR) and reactive (R) closed-loop evaluations on \emph{Val14}, \emph{Test14}, and \emph{Test14-hard}. No post-processing; raw model outputs are evaluated. *: We reproduce the baselines\cite{zheng2025diffusionbased} on the nuPlan\cite{nuplan}, and may differ slightly from the original report, likely due to differences in the randomly sampled 1M training subset. $\dagger$: Motivated by the design of Flow Planner \cite{tan2025flow}, we incorporate Flow Matching and a Mixture-of-Transformer architecture into our Diffusion Forcing Planner.}
\begin{tabular*}{\textwidth}{@{\extracolsep{\fill}}llcccccc@{}}
\toprule
\multirow{2}{*}{\textbf{Type}} & \multirow{2}{*}{\textbf{Planner}}
& \multicolumn{2}{c}{\textbf{Val14}} & \multicolumn{2}{c}{\textbf{Test14}} & \multicolumn{2}{c}{\textbf{Test14-hard}} \\
\cmidrule(lr){3-4}\cmidrule(lr){5-6}\cmidrule(lr){7-8}
& & \textbf{NR} & \textbf{R} & \textbf{NR} & \textbf{R} & \textbf{NR} & \textbf{R} \\
\midrule
\gtext{Expert} & \gtext{Log-replay} & \gtext{93.53} & \gtext{80.32} & \gtext{85.96} & \gtext{68.80} & \gtext{94.03} & \gtext{75.86} \\
\midrule
\multirow{7}{*}{Learning-based}
& PDM-Open*   & 53.53 & 54.24  & 52.81 & 57.23 & 33.51 & 35.83 \\
& UrbanDriver & 68.57 & 64.11 & 51.83 & 67.15 & 50.40 & 49.95 \\
& GameFormer & 13.32 & 8.69 & 11.36 & 9.31 & 7.08 & 6.69 \\
& PlanTF & 84.27 & 76.95 & 85.62 & 79.58 & 69.70 & 61.61 \\
& PLUTO & 88.89 & 78.11 & 89.90 & 78.62 & 70.03 & 59.74 \\
& CoPlanner & 89.48 & 79.00 & 90.31 & 78.81 & 76.82 & {64.47} \\
& Diffusion Planner & 89.87 & 82.80 & 89.19 & 82.93 & 75.99 & 69.22 \\
& Flow Planner & 90.43 & \textbf{83.31} & 89.88 & {82.93} & 76.47 & \textbf{70.42} \\
& Diffusion Planner * & 87.87 & 77.48 & 90.01 & 79.61 & 74.26 & 61.25 \\
& \textbf{DFP (ours)} & {90.33} & {79.97} & \textbf{90.69} & {81.96} & {76.91} & 63.56 \\
& \textbf{DFP (ours) $\dagger$}  & \textbf{92.68} & {81.30} & {90.62} & \textbf{83.59} & \textbf{79.43} & 67.94 \\
\bottomrule
\end{tabular*}
\label{tab:main_results_sup}
\end{table*}

\section{Ablation Study on Temporal Stability: Impact of History Guidance}
\begin{figure*}[htbp]
  \centering
  \begin{subfigure}[t]{0.24\textwidth}
    \centering
    \includegraphics[width=\linewidth]{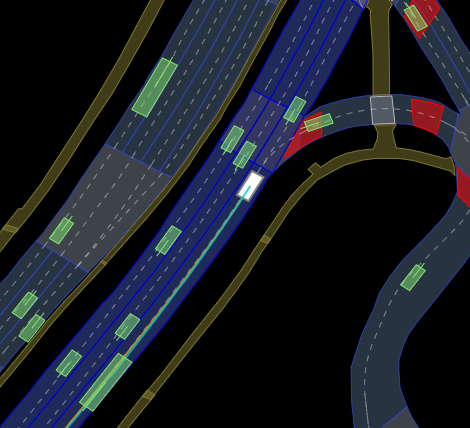}
    \caption{nuPlan scene}
  \end{subfigure}\hfill
  \begin{subfigure}[t]{0.24\textwidth}
    \centering
    \includegraphics[width=\linewidth]{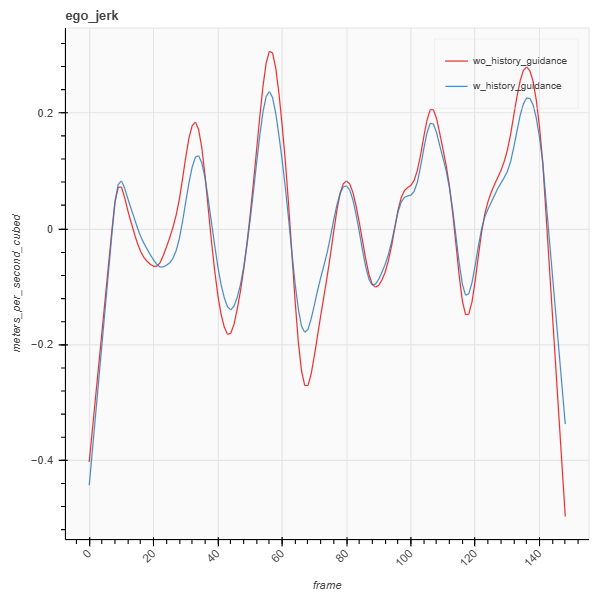}
    \caption{Ego jerk}
  \end{subfigure}\hfill
  \begin{subfigure}[t]{0.24\textwidth}
    \centering
    \includegraphics[width=\linewidth]{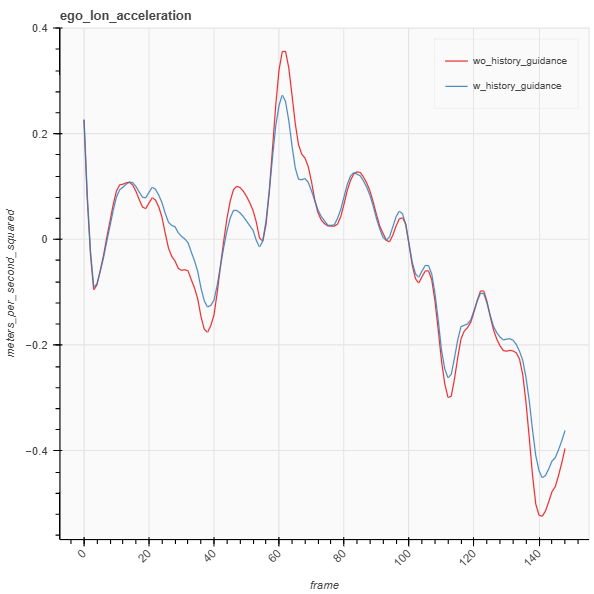}
    \caption{Ego longitudinal acceleration}
  \end{subfigure}\hfill
  \begin{subfigure}[t]{0.24\textwidth}
    \centering
    \includegraphics[width=\linewidth]{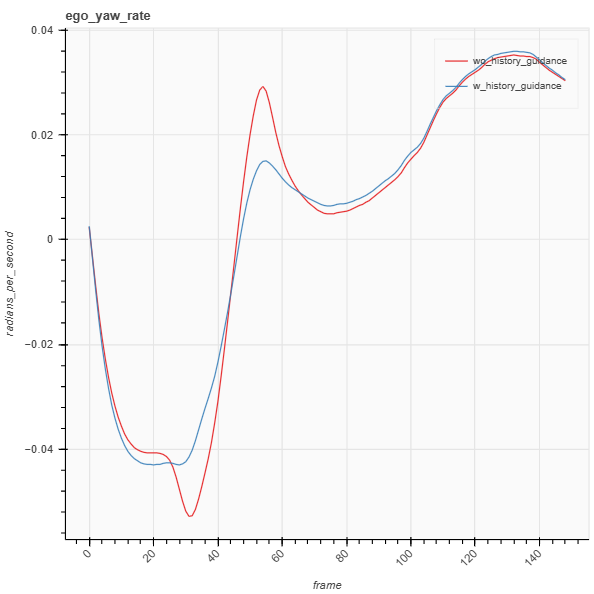}
    \caption{Ego yaw rate}
  \end{subfigure}

  \caption{nuPlan scene visualization and ego kinematic profiles. From left to right: scene view, jerk, longitudinal acceleration, and yaw rate.}
  \label{fig:nuplan_bokeh_kinematics_row}
\end{figure*}

To isolate the contribution of the proposed History-Annealed-Guidance mechanism on temporal stability, we conduct and ablation study comparing the full DFP model (with history guidance) against a variant which predicts history and future trajectories jointly (without history guidance). Figure \ref{fig:nuplan_bokeh_kinematics_row} visualizes the jerk and acceleration over a 15-second segment. The results highlight a clear difference in the quality of the generated trajectories: the model with history guidance generates significantly smoother yaw rate and longitudinal acceleration profiles, indicating reduced jerk and improved comfort.

\begin{figure}[htbp]
  \centering
  \includegraphics[width=0.48\columnwidth]{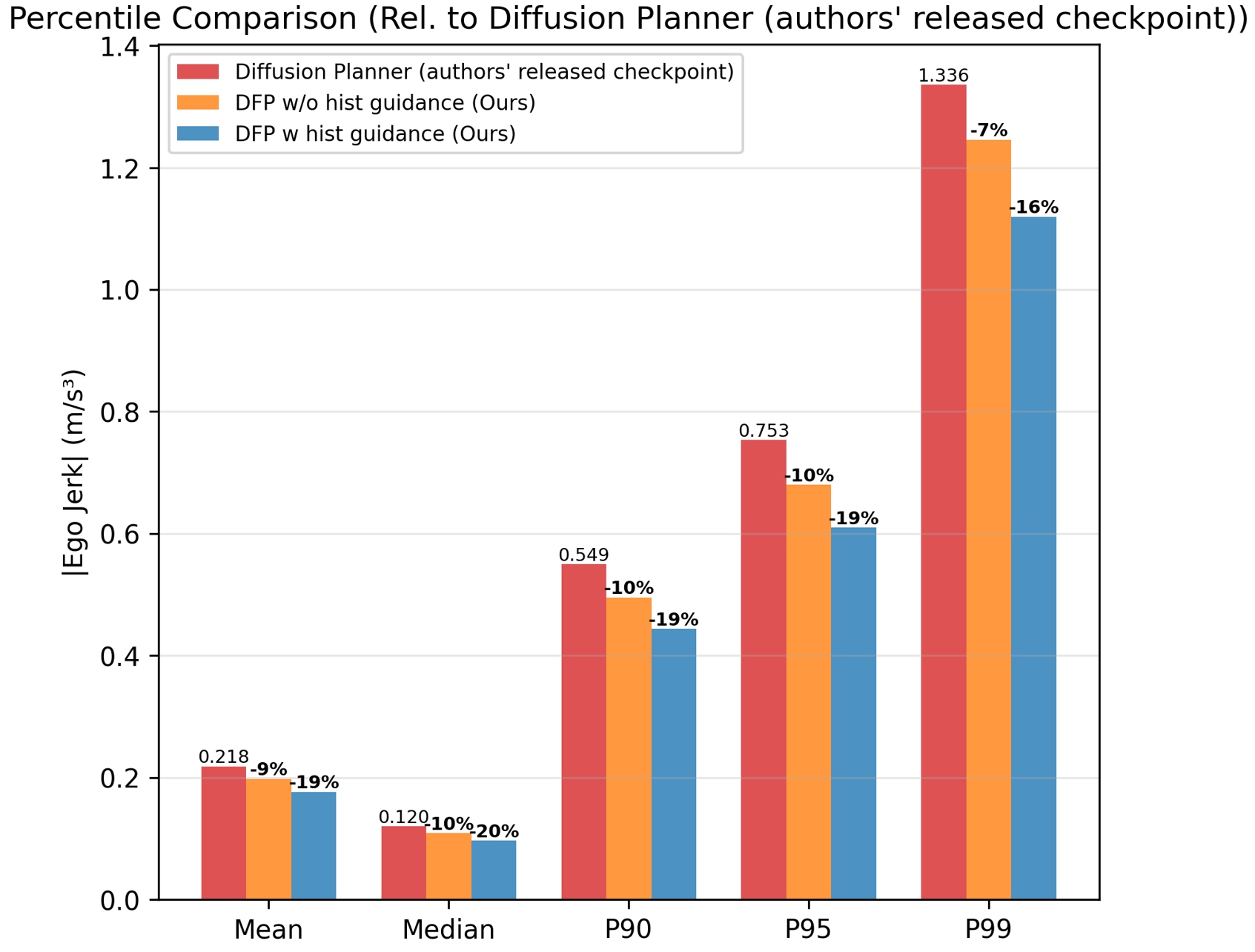}
  \hfill
  \includegraphics[width=0.48\columnwidth]{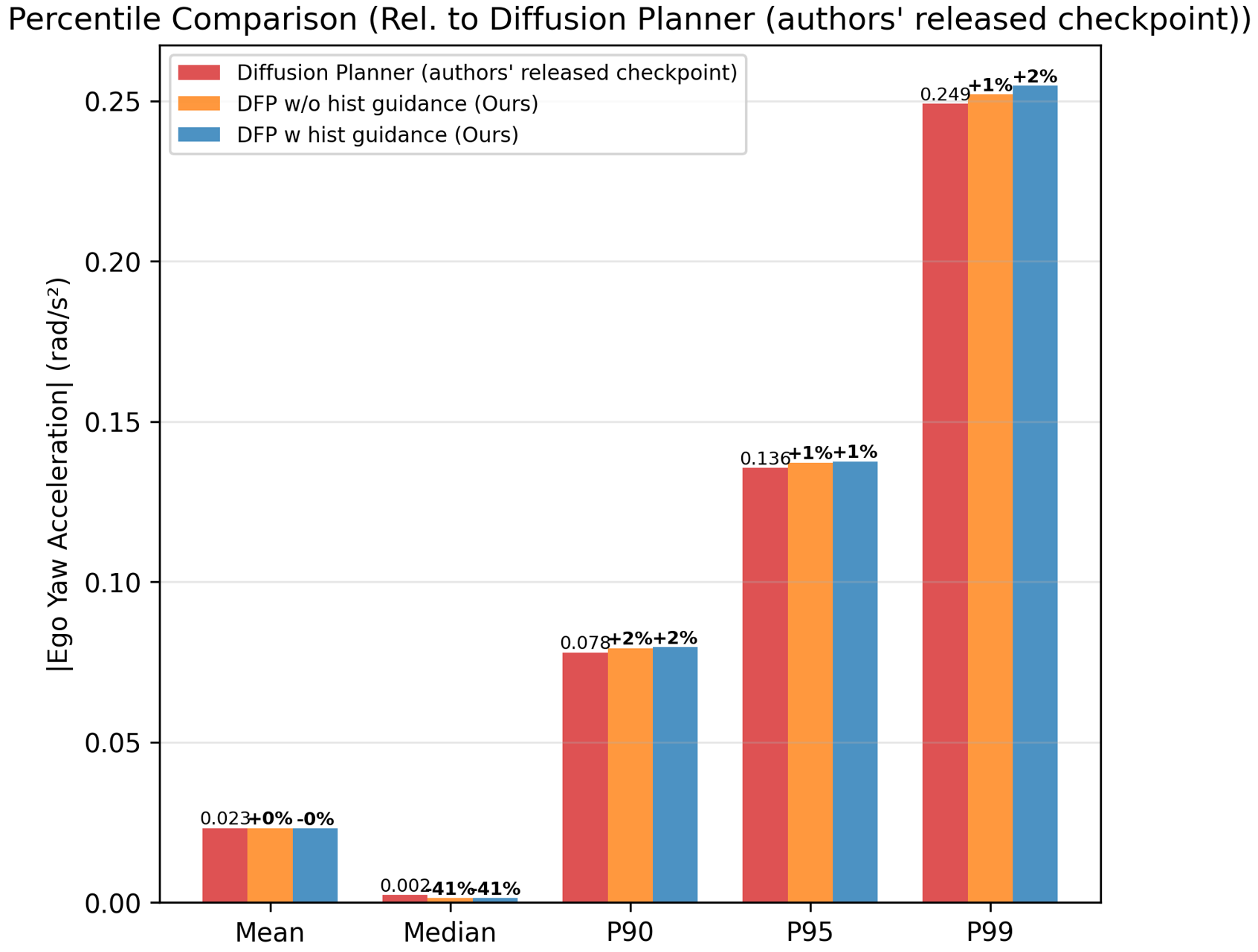}
  \caption{Frame-level comparison.}
  \label{fig:ego_jerk_comparison}
\end{figure}

Figure \ref{fig:ego_jerk_comparison} report the distributions of frame-level quantities. Compared to DP, DFP reduces the mean ego jerk by \textbf{18.9\%} and the standard deviation by \textbf{17.0\%}, indicating not only smoother trajectories on average, but also significantly reduced variability across consecutive planning frames. Notably, on ego yaw acceleration, DFP maintains performance on par with the baseline, aligning with our comfort expectations.
The full CFG inference currently runs at 7.7 FPS (129.79ms), while unguided branch only runs at 14.8 FPS (67.71 ms).

%% file: main.bib
@String(CVPR= {IEEE Conf. Comput. Vis. Pattern Recog.})

@String(CVPR  = {CVPR})

@article{bansal2018chauffeurnet,
  title={Chauffeurnet: Learning to drive by imitating the best and synthesizing the worst},
  author={Bansal, Mayank and Krizhevsky, Alex and Ogale, Abhijit},
  journal={arXiv preprint arXiv:1812.03079},
  year={2018}
}

@article{chen2024end,
  title={End-to-end autonomous driving: Challenges and frontiers},
  author={Chen, Li and Wu, Penghao and Chitta, Kashyap and Jaeger, Bernhard and Geiger, Andreas and Li, Hongyang},
  journal={IEEE Transactions on Pattern Analysis and Machine Intelligence},
  year={2024},
  publisher={IEEE}
}

@article{chitta2022transfuser,
  title={Transfuser: Imitation with transformer-based sensor fusion for autonomous driving},
  author={Chitta, Kashyap and Prakash, Aditya and Jaeger, Bernhard and Yu, Zehao and Renz, Katrin and Geiger, Andreas},
  journal={IEEE transactions on pattern analysis and machine intelligence},
  volume={45},
  number={11},
  pages={12878--12895},
  year={2022},
  publisher={IEEE}
}

@inproceedings{kendall2019learning,
  title={Learning to drive in a day},
  author={Kendall, Alex and Hawke, Jeffrey and Janz, David and Mazur, Przemyslaw and Reda, Daniele and Allen, John-Mark and Lam, Vinh-Dieu and Bewley, Alex and Shah, Amar},
  booktitle={2019 international conference on robotics and automation (ICRA)},
  pages={8248--8254},
  year={2019},
  organization={IEEE}
}

@article{li2024hydra,
  title={Hydra-MDP: End-to-end Multimodal Planning with Multi-target Hydra-Distillation},
  author={Li, Zhenxin and Li, Kailin and Wang, Shihao and Lan, Shiyi and Yu, Zhiding and Ji, Yishen and Li, Zhiqi and Zhu, Ziyue and Kautz, Jan and Wu, Zuxuan and others},
  journal={arXiv preprint arXiv:2406.06978},
  year={2024}
}

@inproceedings{scheel2022urban,
  title={Urban driver: Learning to drive from real-world demonstrations using policy gradients},
  author={Scheel, Oliver and Bergamini, Luca and Wolczyk, Maciej and Osi{\'n}ski, B{\l}a{\.z}ej and Ondruska, Peter},
  booktitle={Conference on Robot Learning},
  pages={718--728},
  year={2022},
  organization={PMLR}
}

@article{hwang2024emma,
  title={Emma: End-to-end multimodal model for autonomous driving},
  author={Hwang, Jyh-Jing and Xu, Runsheng and Lin, Hubert and Hung, Wei-Chih and Ji, Jingwei and Choi, Kristy and Huang, Di and He, Tong and Covington, Paul and Sapp, Benjamin and others},
  journal={arXiv preprint arXiv:2410.23262},
  year={2024}
}

@inproceedings{jiang2023vad,
  title={Vad: Vectorized scene representation for efficient autonomous driving},
  author={Jiang, Bo and Chen, Shaoyu and Xu, Qing and Liao, Bencheng and Chen, Jiajie and Zhou, Helong and Zhang, Qian and Liu, Wenyu and Huang, Chang and Wang, Xinggang},
  booktitle={Proceedings of the IEEE/CVF International Conference on Computer Vision},
  pages={8340--8350},
  year={2023}
}

@article{chen2024vadv2,
  title={Vadv2: End-to-end vectorized autonomous driving via probabilistic planning},
  author={Chen, Shaoyu and Jiang, Bo and Gao, Hao and Liao, Bencheng and Xu, Qing and Zhang, Qian and Huang, Chang and Liu, Wenyu and Wang, Xinggang},
  journal={arXiv preprint arXiv:2402.13243},
  year={2024}
}

@inproceedings{sun2025generalizing,
  title={Generalizing motion planners with mixture of experts for autonomous driving},
  author={Sun, Qiao and Wang, Huimin and Zhan, Jiahao and Nie, Fan and Wen, Xin and Xu, Leimeng and Zhan, Kun and Jia, Peng and Lang, Xianpeng and Zhao, Hang},
  booktitle={2025 IEEE International Conference on Robotics and Automation (ICRA)},
  pages={6033--6039},
  year={2025},
  organization={IEEE}
}

@inproceedings{huang2023gameformer,
  title={Gameformer: Game-theoretic modeling and learning of transformer-based interactive prediction and planning for autonomous driving},
  author={Huang, Zhiyu and Liu, Haochen and Lv, Chen},
  booktitle={Proceedings of the IEEE/CVF International Conference on Computer Vision},
  pages={3903--3913},
  year={2023}
}

@article{cheng2024pluto,
  title={Pluto: Pushing the limit of imitation learning-based planning for autonomous driving},
  author={Cheng, Jie and Chen, Yingbing and Chen, Qifeng},
  journal={arXiv preprint arXiv:2404.14327},
  year={2024}
}

@inproceedings{cheng2024rethinking,
  title={Rethinking imitation-based planners for autonomous driving},
  author={Cheng, Jie and Chen, Yingbing and Mei, Xiaodong and Yang, Bowen and Li, Bo and Liu, Ming},
  booktitle={2024 IEEE International Conference on Robotics and Automation (ICRA)},
  pages={14123--14130},
  year={2024},
  organization={IEEE}
}

@inproceedings{li2024ego,
  title={Is ego status all you need for open-loop end-to-end autonomous driving?},
  author={Li, Zhiqi and Yu, Zhiding and Lan, Shiyi and Li, Jiahan and Kautz, Jan and Lu, Tong and Alvarez, Jose M},
  booktitle={Proceedings of the IEEE/CVF Conference on Computer Vision and Pattern Recognition},
  pages={14864--14873},
  year={2024}
}

@article{ho2020denoising,
  title={Denoising diffusion probabilistic models},
  author={Ho, Jonathan and Jain, Ajay and Abbeel, Pieter},
  journal={Advances in neural information processing systems},
  volume={33},
  pages={6840--6851},
  year={2020}
}

@inproceedings{janner2022diffuser,
  title = {Planning with Diffusion for Flexible Behavior Synthesis},
  author = {Michael Janner and Yilun Du and Joshua B. Tenenbaum and Sergey Levine},
  booktitle = {International Conference on Machine Learning},
  year = {2022},
}

@inproceedings{
    ajay2023is,
    title={Is Conditional Generative Modeling all you need for Decision Making?},
    author={Anurag Ajay and Yilun Du and Abhi Gupta and Joshua B. Tenenbaum and Tommi S. Jaakkola and Pulkit Agrawal},
    booktitle={The Eleventh International Conference on Learning Representations },
    year={2023},
    url={https://openreview.net/forum?id=sP1fo2K9DFG}
}

@article{chi2025diffusion,
  title={Diffusion policy: Visuomotor policy learning via action diffusion},
  author={Chi, Cheng and Xu, Zhenjia and Feng, Siyuan and Cousineau, Eric and Du, Yilun and Burchfiel, Benjamin and Tedrake, Russ and Song, Shuran},
  journal={The International Journal of Robotics Research},
  volume={44},
  number={10-11},
  pages={1684--1704},
  year={2025},
  publisher={Sage Publications Sage UK: London, England}
}

@inproceedings{
tan2025flow,
title={Flow Matching-Based Autonomous Driving Planning with Advanced Interactive Behavior Modeling},
author={Tianyi Tan and Yinan Zheng and Ruiming Liang and Zexu Wang and Kexin Zheng and Jinliang Zheng and Jianxiong Li and Xianyuan Zhan and Jingjing Liu},
booktitle={The Thirty-ninth Annual Conference on Neural Information Processing Systems},
year={2025}
}

@inproceedings{
zheng2025diffusionbased,
title={Diffusion-Based Planning for Autonomous Driving with Flexible Guidance},
author={Yinan Zheng and Ruiming Liang and Kexin ZHENG and Jinliang Zheng and Liyuan Mao and Jianxiong Li and Weihao Gu and Rui Ai and Shengbo Eben Li and Xianyuan Zhan and Jingjing Liu},
booktitle={The Thirteenth International Conference on Learning Representations},
year={2025}
}

@article{diffusiondrive,
  title={DiffusionDrive: Truncated Diffusion Model for End-to-End Autonomous Driving},
  author={Bencheng Liao and Shaoyu Chen and Haoran Yin and Bo Jiang and Cheng Wang and Sixu Yan and Xinbang Zhang and Xiangyu Li and Ying Zhang and Qian Zhang and Xinggang Wang},
  journal={arXiv preprint arXiv:2411.15139},
  year={2024}
}

@inproceedings{xing2025goalflow,
  title={Goalflow: Goal-driven flow matching for multimodal trajectories generation in end-to-end autonomous driving},
  author={Xing, Zebin and Zhang, Xingyu and Hu, Yang and Jiang, Bo and He, Tong and Zhang, Qian and Long, Xiaoxiao and Yin, Wei},
  booktitle={Proceedings of the Computer Vision and Pattern Recognition Conference},
  pages={1602--1611},
  year={2025}
}

@article{song2025history,
  title={History-guided video diffusion},
  author={Song, Kiwhan and Chen, Boyuan and Simchowitz, Max and Du, Yilun and Tedrake, Russ and Sitzmann, Vincent},
  journal={arXiv preprint arXiv:2502.06764},
  year={2025}
}

@article{black2025real,
  title={Real-Time Execution of Action Chunking Flow Policies},
  author={Black, Kevin and Galliker, Manuel Y and Levine, Sergey},
  journal={arXiv preprint arXiv:2506.07339},
  year={2025}
}

@inproceedings{jiang2025streaming,
  title     = {Streaming Flow Policy: Simplifying diffusion/flow-matching policies by treating action trajectories as flow trajectories},
  author    = {Sunshine Jiang AND Xiaolin Fang AND Nicholas Roy AND Tom{\'a}s Lozano-P{\'e}rez AND Leslie Pack Kaelbling AND Siddharth Ancha},
  booktitle = {9th Annual Conference on Robot Learning, CoRL 2025},
  year      = {2025},
  address   = {Seoul, Korea},
  month     = {September},
  url       = {https://openreview.net/forum?id=jnpILGz9gQ}
}

@inproceedings{liu2024bid,
  title={Bidirectional Decoding: Improving Action Chunking via Guided Test-Time Sampling},
  author={Yuejiang Liu and Jubayer Ibn Hamid and Annie Xie and Yoonho Lee and Max Du and Chelsea Finn},
  booktitle={The Thirteenth International Conference on Learning Representations},
  year={2025},
  url={https://openreview.net/forum?id=qZmn2hkuzw}
}

@misc{torne2025learninglongcontextdiffusionpolicies,
      title={Learning Long-Context Diffusion Policies via Past-Token Prediction}, 
      author={Marcel Torne and Andy Tang and Yuejiang Liu and Chelsea Finn},
      year={2025},
      eprint={2505.09561},
      archivePrefix={arXiv},
      primaryClass={cs.RO},
      url={https://arxiv.org/abs/2505.09561}, 
}

@article{ho2022classifier,
  title={Classifier-free diffusion guidance},
  author={Ho, Jonathan and Salimans, Tim},
  journal={arXiv preprint arXiv:2207.12598},
  year={2022}
}

@article{de2019causal,
  title={Causal confusion in imitation learning},
  author={De Haan, Pim and Jayaraman, Dinesh and Levine, Sergey},
  journal={Advances in neural information processing systems},
  volume={32},
  year={2019}
}

@article{
  liang2024mixture,
  title={Mixture-of-Transformers: A Sparse and Scalable Architecture for Multi-Modal Foundation Models},
  author={Weixin Liang and LILI YU and Liang Luo and Srini Iyer and Ning Dong and Chunting Zhou and Gargi Ghosh and Mike Lewis and Wen-tau Yih and Luke Zettlemoyer and Xi Victoria Lin},
  journal={Transactions on Machine Learning Research},
  issn={2835-8856},
  year={2025},
  url={https://openreview.net/forum?id=Nu6N69i8SB},
  note={}
}

@INPROCEEDINGS{nuplan, 
  title={NuPlan: A closed-loop ML-based planning benchmark for autonomous vehicles},
  author={Caesar, Holger and Kabzan, Juraj and Tan, Kok Seang and Fong, Whye Kit and Wolff, Eric and Lang, Alex and Fletcher, Luke and Beijbom, Oscar and Omari, Sammy},
  booktitle={CVPR ADP3 workshop},
  year={2021}
}

@article{black2024pi_0,
  title={$\pi_0 $: A Vision-Language-Action Flow Model for General Robot Control},
  author={Black, Kevin and Brown, Noah and Driess, Danny and Esmail, Adnan and Equi, Michael and Finn, Chelsea and Fusai, Niccolo and Groom, Lachy and Hausman, Karol and Ichter, Brian and others},
  journal={arXiv preprint arXiv:2410.24164},
  year={2024}
}

@article{zhao2025diffe2e,
  title={Diffe2e: Rethinking end-to-end driving with a hybrid action diffusion and supervised policy},
  author={Zhao, Rui and Fan, Yuze and Chen, Ziguo and Gao, Fei and Gao, Zhenhai},
  journal={arXiv preprint arXiv:2505.19516},
  year={2025}
}

@article{wen2024diffusion,
  title={Diffusion-vla: Generalizable and interpretable robot foundation model via self-generated reasoning},
  author={Wen, Junjie and Zhu, Minjie and Zhu, Yichen and Tang, Zhibin and Li, Jinming and Zhou, Zhongyi and Li, Chengmeng and Liu, Xiaoyu and Peng, Yaxin and Shen, Chaomin and others},
  journal={arXiv preprint arXiv:2412.03293},
  year={2024}
}

@inproceedings{black2025_pi05,
  title={$\pi_0 $: A Vision-Language-Action Model with Open-World Generalization},
  author={Black, Kevin and Brown, Noah and Darpinian, James and Dhabalia, Karan and Driess, Danny and Esmail, Adnan and Equi, Michael Robert and Finn, Chelsea and Fusai, Niccolo and Galliker, Manuel Y and others},
  booktitle={9th Annual Conference on Robot Learning},
  year={2025}
}

@article{mandlekar2021matters,
  title={What matters in learning from offline human demonstrations for robot manipulation},
  author={Mandlekar, Ajay and Xu, Danfei and Wong, Josiah and Nasiriany, Soroush and Wang, Chen and Kulkarni, Rohun and Fei-Fei, Li and Savarese, Silvio and Zhu, Yuke and Mart{\'\i}n-Mart{\'\i}n, Roberto},
  journal={arXiv preprint arXiv:2108.03298},
  year={2021}
}

@inproceedings{chen2025drivinggpt,
  title={Drivinggpt: Unifying driving world modeling and planning with multi-modal autoregressive transformers},
  author={Chen, Yuntao and Wang, Yuqi and Zhang, Zhaoxiang},
  booktitle={Proceedings of the IEEE/CVF International Conference on Computer Vision},
  pages={26890--26900},
  year={2025}
}

@article{song2020score,
  title={Score-based generative modeling through stochastic differential equations},
  author={Song, Yang and Sohl-Dickstein, Jascha and Kingma, Diederik P and Kumar, Abhishek and Ermon, Stefano and Poole, Ben},
  journal={arXiv preprint arXiv:2011.13456},
  year={2020}
}

@article{caesar2021nuplan,
  title={nuplan: A closed-loop ml-based planning benchmark for autonomous vehicles},
  author={Caesar, Holger and Kabzan, Juraj and Tan, Kok Seang and Fong, Whye Kit and Wolff, Eric and Lang, Alex and Fletcher, Luke and Beijbom, Oscar and Omari, Sammy},
  journal={arXiv preprint arXiv:2106.11810},
  year={2021}
}

@article{kesting2013traffic,
  title={Traffic flow dynamics: data, models and simulation},
  author={Kesting, Arne and Treiber, Martin},
  journal={No. Book, Whole)(Springer Berlin Heidelberg, Berlin, Heidelberg},
  year={2013},
  publisher={Springer}
}

@inproceedings{dauner2023parting,
  title={Parting with misconceptions about learning-based vehicle motion planning},
  author={Dauner, Daniel and Hallgarten, Marcel and Geiger, Andreas and Chitta, Kashyap},
  booktitle={Conference on Robot Learning},
  pages={1268--1281},
  year={2023},
  organization={PMLR}
}

@article{zhong2025coplanner,
  title={CoPlanner: An Interactive Motion Planner with Contingency-Aware Diffusion for Autonomous Driving},
  author={Zhong, Ruiguo and Yao, Ruoyu and Liu, Pei and Chen, Xiaolong and Yang, Rui and Ma, Jun},
  journal={arXiv preprint arXiv:2509.17080},
  year={2025}
}

@article{lu2022dpm,
  title={Dpm-solver: A fast ode solver for diffusion probabilistic model sampling in around 10 steps},
  author={Lu, Cheng and Zhou, Yuhao and Bao, Fan and Chen, Jianfei and Li, Chongxuan and Zhu, Jun},
  journal={Advances in neural information processing systems},
  volume={35},
  pages={5775--5787},
  year={2022}
}
